\documentclass[10pt]{article}

\usepackage[a4paper,top=0.95in,bottom=0.95in,left=0.85in,right=0.85in]{geometry}
\usepackage[T1]{fontenc}
\usepackage{lmodern}
\usepackage{microtype}
\usepackage{amsmath,amssymb}
\usepackage{graphicx}
\usepackage{booktabs}
\usepackage{array}
\usepackage{xcolor}
\usepackage{caption}
\usepackage{subcaption}
\usepackage{float}
\usepackage{placeins}
\usepackage{enumitem}
\usepackage{titlesec}
\usepackage[numbers,sort&compress]{natbib}
\usepackage{hyperref}
\hypersetup{colorlinks=true,linkcolor=blue!55!black,citecolor=blue!55!black,urlcolor=blue!55!black,
  pdftitle={Double-Helix Active Geometry: LiDAR-Anchored Multi-View Depth with Selective Abstention},
  pdfauthor={Jinwen Wen}}
\titlespacing*{\section}{0pt}{1.05em}{0.4em}
\titlespacing*{\subsection}{0pt}{0.7em}{0.3em}
\renewcommand{\dh}{DH\nobreakdash-Active}

\title{\vspace{-1.2em}\textbf{Double-Helix Active Geometry:\\[2pt]
LiDAR-Anchored Multi-View Depth with Selective Abstention}}

\author{
  Jinwen Wen\\
  Independent Researcher\\
 }
\date{\today}

\begin{document}
\maketitle

\begin{abstract}
\noindent
Consumer depth sensors such as the LiDAR scanner on recent iPhones provide metric range, but
their useful range is short and their returns are sparse. We present \textbf{\dh{}}, a
lightweight, training-free geometry back-end that treats the sensor as a \emph{metric ruler}
rather than the sole source of depth. Near-field returns anchor the metric relative pose of two
views through PnP; visually trackable samples without a valid depth return are then triangulated
under that pose. A parallax/reprojection gate abstains wherever the geometry is ill-conditioned,
leaving an explicit hole and a selective score instead of forcing an estimate. The measured core
front end---spiral sampling, sparse back-projection, and hole taxonomy, excluding preprocessing
and multi-view recovery---runs at \textbf{1.11\,ms} median latency on CPU (OpenCV using 14
threads), about $38\times$ faster than a DINOv2-L visual branch on GPU in our timing setup.
Across two iPhone captures and the public TUM RGB-D and ARKitScenes benchmarks, held-out depth is
recovered at \mbox{1.4--6.7\%} median relative error. In a controlled ARKitScenes protocol that
uses only returns within 2\,m to set scale and an independent laser scan as ground truth,
\dh{} achieves \textbf{64.2\% scene-median coverage} of evaluable far-field candidates at
\textbf{13.4\% scene-median relative error}; direct triangulation from the device trajectory is
not usable. We also report the alternatives that failed in our tests: single-frame defocus,
classical focus-stack depth, defocus--LiDAR fusion, point-to-point ICP over a good
visual-inertial track, and attention-to-holes resampling. A 1.26\,B learned model remains more
accurate after oracle scale alignment. The contribution here is narrower: metric sparse depth,
explicit abstention, zero learned parameters, and near-millisecond CPU cost.
\end{abstract}

\section{Introduction}

A robot or embodied agent that wants to act in a room needs to know where surfaces are, in
metres, and---just as importantly---needs to know \emph{where it does not know}. Active depth
sensors such as the LiDAR scanner on recent iPhones are metric and robust, but their usable
range is short: distant, reflective, or grazing surfaces often receive no return. Learned depth
completion and multi-view systems can fill gaps, but they require training and usually a GPU for
practical throughput. Large visual encoders used in vision--language--action (VLA)
systems~\citep{kim2024openvla,oquab2024dinov2} provide powerful image features, not by themselves
metric depth. This leaves room for a small geometric back-end that uses the metric signal already
on the device and reports when it has too little evidence.

This paper asks a narrow, deployable question: \emph{on a device that already carries a
short-range metric sensor, how far can purely geometric reasoning extend that sensor, without
any learning, while remaining honest about its own failures?} Our answer is a system,
\dh{}, organised around a single principle:

\begin{quote}
\itshape Demote the strong-but-short-range active sensor from the only depth source to a
\textbf{metric ruler}: use its near returns to fix the metric relative pose between views, use
that pose to triangulate the far field it cannot itself reach, leave an explicit hole where
neither can see, and attach a selective confidence to every estimate.
\end{quote}

Concretely, \dh{} samples each frame along a pair of foveated spiral
trajectories~\citep{schwartz1977spatial,traver2010review}, back-projects the near LiDAR
returns at those samples into a confidence-tagged sparse 3D set, and classifies every
\emph{missing} sample by why it is missing. For samples that lack a valid depth return but
remain visually trackable, it estimates the metric two-view relative pose by solving
\mbox{Perspective-n-Point}~\citep{lepetit2009epnp,fischler1981ransac} on the near points that
\emph{do} have depth, then triangulates the far points under that pose. A parallax and
reprojection gate discards ill-conditioned rays, so the system reports ``could not recover''
rather than a confident guess---a deliberate design choice in the spirit of selective
prediction~\citep{geifman2017selective,elyaniv2010foundations}.

We stress at the outset what this work is and is not. It is \emph{not} a new state of the art
in depth accuracy: a 1.26\,B-parameter learned multi-view model is markedly more accurate,
albeit only up to scale and only on a GPU (\S\ref{sec:baselines}). It \emph{is} a study of how
much honest, metric geometry one can extract on commodity hardware at near-zero cost, and of
where that approach breaks. The word ``active'' in the name refers to the active depth input and
the foveated-sampling line
of work from which the project grew. The evaluated system does not plan camera motion or choose
robot actions. Our contributions are:

\begin{enumerate}[leftmargin=1.35em,topsep=2pt,itemsep=1.5pt]
\item \textbf{A LiDAR-as-ruler far-field depth pipeline.} A self-consistent loop in which the
same near-field returns supply both metric depth and the pose scale needed to triangulate the
far field, recovering points genuinely beyond sensor range, with no network and no training
(\S\ref{sec:method}).
\item \textbf{Abstention as a first-class output.} A parallax/reprojection gate and a per-point
selective score that trade coverage for tail-risk; we show on a strict near/far protocol that
the gate, not the resampling, is the core safety mechanism (\S\ref{sec:e8},\,\ref{sec:ablation}).
\item \textbf{A controlled-cutoff evaluation with independent ground truth} that directly
measures far-field recovery on a fixed evaluable cohort: all returns beyond 2\,m are masked and
the outputs are scored against a static laser scan (\S\ref{sec:e8}).
\item \textbf{A documented set of negative results}---defocus depth, focus stacks, defocus--LiDAR
fusion, ICP over a good visual-inertial track, and attention-to-holes---tested with explicit
controls (\S\ref{sec:negative}). These delimit where the simple geometry is and is not the right
tool, and are, we argue, as useful as the positive results.
\item \textbf{An efficiency characterisation} against a DINOv2-L visual branch on the same machine:
$\sim$38$\times$ lower measured latency for the core CPU front end versus the encoder on a GPU,
at zero learned parameters (\S\ref{sec:latency}), plus a descriptive downstream comparison
(\S\ref{sec:downstream}).
\end{enumerate}

\section{Related Work}

\paragraph{Structure from motion and SLAM.}
Recovering pose and structure from multiple views is classical~\citep{hartley2004multiple}.
Modern pipelines---COLMAP for offline reconstruction~\citep{schonberger2016colmap},
ORB-SLAM3 for real-time tracking~\citep{campos2021orbslam3}---estimate camera motion and a
(typically dense or semi-dense) map jointly. Metric scale comes from cues such as calibrated
stereo, RGB-D, or inertial measurements; loop closure corrects drift but does not itself supply
absolute scale. \dh{} deliberately occupies a smaller niche: it does not
build or optimise a global map, performs no bundle adjustment or loop closure, and instead uses
an already-present metric sensor to fix two-view scale locally. Where RGB-D fusion methods such
as KinectFusion~\citep{newcombe2011kinectfusion} integrate \emph{dense} depth into a
volumetric field~\citep{curless1996volumetric}, our measurements are sparse spiral samples and
our emphasis is on the points the depth sensor \emph{misses}.

\paragraph{Sparse depth completion.}
RGB-D completion commonly densifies a sparse or incomplete depth map with learned image
priors~\citep{zhang2018deepdepthcompletion,wong2019unsupervised}. Recent systems also combine
sparse direct time-of-flight returns with dense image features~\citep{kim2026densemetric} or use
LiDAR to prompt spatio-temporal multi-view stereo~\citep{sun2026drivemvs}. These methods target
dense prediction and generally rely on learned components, supplied poses, or both. \dh{} makes
a different trade: it produces only sparse metric estimates, learns no image prior, estimates
the two-view metric pose from the near returns themselves, and abstains when the recovered ray
geometry is weak.

\paragraph{Pose from known-depth points.}
Solving for camera pose from 2D--3D correspondences is the Perspective-n-Point
problem~\citep{lepetit2009epnp}, robustified with RANSAC~\citep{fischler1981ransac}. Using
LiDAR-backed points to anchor metric pose, and short Lucas--Kanade~\citep{lucas1981iterative}
tracks of Shi--Tomasi features~\citep{shi1994good} to form correspondences, are individually
standard. Our contribution is not a new estimator but the \emph{closed loop}: the near returns
that the sensor provides for free are exactly what make the far-field triangulation metric, and
we show that substituting the device's own visual-inertial pose for this PnP step collapses the
result (\S\ref{sec:e8}).

\paragraph{Learned multi-view geometry.}
Recent transformers regress geometry directly from images: DUSt3R aligns image pairs into a
common point map~\citep{wang2024dust3r}, and VGGT predicts cameras, depth, and point maps
feed-forward~\citep{wang2025vggt}. These are accurate but output structure \emph{up to scale},
carry hundreds of millions to billions of parameters, and are normally run on a GPU for
practical throughput. We use them as reference points (\S\ref{sec:baselines}); the largest is
more accurate after scale alignment, whereas \dh{} obtains metric scale from the sensor rather
than a ground-truth fit.

\paragraph{Passive depth cues.}
Depth from defocus and from focus stacks infers range from the spatial variation of image
sharpness using hand-designed focus measures~\citep{pertuz2013analysis}. These need no emitter
and no training, which makes them attractive on the same grounds as our method; in
\S\ref{sec:negative} we report where the classical estimators fail in our in-house and public
data tests, and explain why.

\paragraph{Foveated sampling.}
Space-variant, centre-dense sampling has a long history in biological and robotic
vision~\citep{schwartz1977spatial,strasburger2011peripheral,traver2010review}, and spiral and
phyllotactic arrangements tile the plane with a graded local density~\citep{vogel1979better,
cook1986stochastic}. \dh{} uses a dual spiral purely as a cheap, training-free sampling
template that concentrates measurements near the gaze centre; we make no claim that the spiral
itself recovers depth (a claim we explicitly falsify in \S\ref{sec:negative}).

\paragraph{Uncertainty and selective prediction.}
Reporting calibrated confidence~\citep{guo2017calibration} and allowing a model to abstain on
inputs it cannot handle~\citep{geifman2017selective,elyaniv2010foundations} are well studied for
classifiers. We bring the \emph{abstention} stance to a geometry back-end: the gate is a reject
option over rays, and the per-point score induces a risk--coverage trade-off. Consistent with
our own measurements (\S\ref{sec:crossdata},\,\ref{sec:e8}), we are careful to call this score
\emph{selective}, not calibrated.

\section{Method}
\label{sec:method}

\dh{} processes a short monocular RGB stream with a registered, sparse, short-range depth
channel (the iPhone main camera and LiDAR, exposed through Record3D). The pipeline has four
stages; none contains a neural network.

\begin{figure}[H]
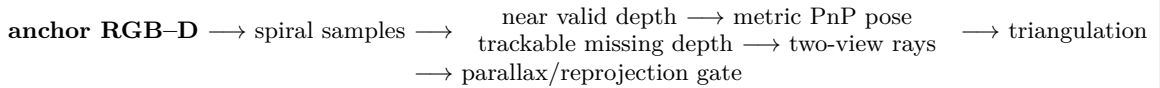

\centering
\fbox{\begin{minipage}{0.91\textwidth}
\centering\small
\textbf{anchor RGB--D} $\longrightarrow$ spiral samples $\longrightarrow$
\begin{tabular}{c}
near valid depth $\longrightarrow$ metric PnP pose\\[-1pt]
trackable missing depth $\longrightarrow$ two-view rays
\end{tabular}
$\longrightarrow$ triangulation $\longrightarrow$ parallax/reprojection gate
\end{minipage}}
\caption{The metric loop. Near returns determine the two-view pose and scale; missing but
trackable samples provide the rays to triangulate. The last gate either keeps a metric estimate
with a selective score or leaves the sample empty.}
\label{fig:pipeline}
\end{figure}

\subsection{Spiral sampling and confidence-aware sparse 3D}
Each frame is sampled at $N$ points along a pair of phase-shifted spiral trajectories whose
radial density is highest at the gaze centre and decays toward the periphery, following the
foveated-sampling tradition~\citep{schwartz1977spatial,traver2010review}. At each sample with a
valid depth return $z$ and intrinsics $(f_x,f_y,c_x,c_y)$ we back-project to a camera-frame
point $\big(\tfrac{(u-c_x)z}{f_x},\,\tfrac{(v-c_y)z}{f_y},\,z\big)$, and tag it with a confidence
and a set of flags (\texttt{NO\_DEPTH}, \texttt{LOW\_CONF}, \texttt{SPECULAR},
\texttt{SATURATED}, \texttt{UNDEREXPOSED}). Samples with no usable return are set to NaN and
flagged---``fail loud, fail localized''---rather than interpolated.

\paragraph{Hole taxonomy.}
A sample can be missing for very different reasons, and the right response differs. We classify
each empty sample using its appearance flags plus a trackability score (the Shi--Tomasi minimum
eigenvalue~\citep{shi1994good} at that pixel). The key class is \texttt{TRACKABLE\_NO\_LIDAR}:
normal brightness and sufficient texture, but no usable depth. It is a candidate for multi-view
recovery, not a claim about why the return is absent: range, low confidence, occlusion boundaries,
and reflectance can all produce the same label. On our two captures this class accounts for
\textbf{90.4\%} (indoor) and \textbf{86.1\%} (outdoor) of all holes, and its trackability is of
the same order as that of valid-depth samples. Thus most observed holes still provide an RGB
signal that can be tested by triangulation.

\subsection{LiDAR-anchored relative pose}
\label{sec:pose}
To triangulate, we need the metric relative pose between two frames. The device supplies a
visual-inertial (VIO) pose, but we find it unusable for triangulation: its rotation carries a
$\sim$5$^\circ$ systematic offset (VIO/RGB desynchronisation and drift), and feeding it in yields
depth correlated with ground truth at only $\rho\!\approx\!0.16$. Instead, for each keyframe pair
we self-calibrate the metric relative pose by solving PnP~\citep{lepetit2009epnp} with
RANSAC~\citep{fischler1981ransac} on the \emph{near} points that have LiDAR depth: the sensor
provides both the anchoring 3D coordinates and the absolute scale, in one self-consistent loop.
Correspondences are short Lucas--Kanade tracks~\citep{lucas1981iterative} of spiral samples.

\subsection{Gated two-view triangulation and abstention}
\label{sec:gate}
Given the anchored pose, each far-field correspondence is triangulated~\citep{hartley2004multiple}.
We then \emph{abstain} on any estimate that is geometrically ill-conditioned, keeping a point only
if (i) the triangulation parallax exceeds $0.3^\circ$ and (ii) the reprojection residual is below
$3$\,px. Endpoints may use distinct intrinsics ($K_\text{anchor}\!\neq\!K_\text{current}$), which
we verify with a synthetic regression test. The gate is the system's reject option: it is what
turns an over-confident far-field guess into an explicit, reported hole.

\subsection{Selective confidence}
Each surviving point carries a continuous score derived from its parallax and reprojection
residual. Sweeping a threshold over this score induces a risk--coverage curve, and we use it as a
\emph{selective} signal. Higher thresholds keep fewer points and initially reduce error, but the
trend is not strictly monotone. We show in \S\ref{sec:crossdata} and \S\ref{sec:e8} that the
score provides a useful ordering while its expected calibration error remains too high to call
it a calibrated probability; we therefore avoid that word throughout.

\subsection{Global integration (scope note)}
Recovered points can be fused into a confidence-weighted truncated signed-distance
field~\citep{curless1996volumetric,newcombe2011kinectfusion} for visualisation, with overlap
regions blended by confidence rather than overwritten. This widens map coverage but, as we report
honestly in \S\ref{sec:limitations}, does \emph{not} stitch the sparse single-view fragments into
a closed room; global mapping is outside the scope of the claims we defend here.

\section{What Does Not Work}
\label{sec:negative}

The design above is the residue of a sequence of alternatives that did not survive their
screening tests. Most use the in-house captures; the focus-stack study instead uses the public
synthetic FoD500 set~\citep{maximov2020focus} and real DDFF-12 light-field
data~\citep{hazirbas2018deep}. We report these results because they delimit where simple geometry
is the wrong tool and explain why the final system relies on the ruler-and-gate loop.

\begin{itemize}[leftmargin=1.2em,topsep=2pt,itemsep=2pt]
\item \textbf{Dual-helix temporal cross-correlation.} The project's original intuition---that two
phase-shifted 1D spiral signals could be cross-correlated to ``squeeze out'' parallax---is
\emph{false}. Against LiDAR inverse depth it scores $r\!=\!-0.04$, versus $+0.34$ for plain 2D
optical flow: under forward motion the flow is radial and a 1D tangential correlation is
structurally blind to it.
\item \textbf{Single-frame depth from defocus.} The defocus signal is real but weak ($|\rho|$
median $0.22$) and dominated by two ambiguities: texture (a blank wall looks defocused) and sign
(near vs.\ far of focus). A sign-only oracle reaches $r\!=\!0.66$, but a deployable single-frame
estimator that must \emph{produce} the sign collapses to $r\!=\!0.06$.
\item \textbf{Classical focus-stack depth.} On clean synthetic data argmax-sharpness reaches
Spearman $0.69$ (gated Pearson $0.76$), but on real light-field data it swings with the scene,
flips sign on $\sim$1/5 of scenes, and confidence gating cannot rescue it (it is
``confidently wrong''): top-25\% $|r|$ rises only $0.15\!\to\!0.21$.
\item \textbf{Defocus--LiDAR fusion for densification.} On this hardware plain bilinear upsampling
of LiDAR already reaches $r\!\approx\!0.82$--$0.99$; defocus guidance never beats it and slightly
\emph{raises} error. The sensor is too clean to leave defocus any detail to add.
\item \textbf{Point-to-point ICP over a good VIO track.} With a reliable device pose, ICP between
two independently resampled sparse sets is ill-posed and rarely confident; trusting VIO alone
covers \emph{more} (6{,}200 vs.\ 5{,}417 voxels) than applying ICP. ICP earns its keep only when
no reliable pose exists. This is the same lesson as defocus-vs-LiDAR: a strong prior makes the
clever add-on redundant.
\item \textbf{Attention-to-holes resampling.} Moving the foveal centre onto the largest hole
cluster makes coverage \emph{worse} (hole rate $0.67\!\to\!0.78$): the holes here are sensor
missing-depth regions rather than a shortage of sample locations, so resampling cannot create a
return the sensor never gave. This result supports using the active layer for \emph{routing}
(``this may need a head turn, not an eye movement'') rather than claiming that it fills depth.
\end{itemize}

\section{Experimental Setup}
\label{sec:setup}

\paragraph{Data.} We use two in-house iPhone captures via Record3D (one indoor, near-static
hand-held; one outdoor, walking), and public datasets for the positive and negative studies:
\textbf{TUM RGB-D} (11 sequences,
structured-light depth)~\citep{sturm2012tum}; \textbf{ARKitScenes} (Apple-LiDAR RGB-D with a
registered high-resolution laser scan)~\citep{baruch2021arkitscenes}; and \textbf{RH20T} (real
RGB-D with robot action labels)~\citep{fang2023rh20t}. The focal-stack screening uses
\textbf{FoD500} (500 synthetic five-image stacks)~\citep{maximov2020focus} and
\textbf{DDFF-12} (real refocused light-field stacks from 12 scenes)~\citep{hazirbas2018deep}.
Public datasets are obtained from their official sources; no data or model weights are
redistributed.

\paragraph{Protocol.} Our central metric is the relative error of recovered metric depth against
a reference, reported as a per-sequence median and then a scene-median over sequences (to keep
long videos from dominating), alongside the 90th percentile (p90) as a tail-risk measure, Pearson
correlation, and \emph{coverage} (fraction of evaluable candidates the method chooses to output).
Reporting coverage beside accuracy is essential for a method that abstains. Unless noted, depth
hold-outs split points so that pose calibration and evaluation never share a point.

\paragraph{Reproducibility.} The geometry runtime has no deep-learning dependencies; 72 offline
unit tests cover the pipeline. Each experiment archives a JSON of its full configuration and
per-sequence numbers. Baseline encoders (DINOv2, DUSt3R, VGGT) use official code and checkpoints
fetched separately and are timed/scored as-is.

\section{Results}

\subsection{Deployment efficiency}
\label{sec:latency}
We time the core \dh{} front-end stages against the DINOv2-L visual branch of a VLA encoder on
the same Apple M4 Pro (batch 1, fp32, warm-up, $p50$/$p95$/$p99$; DINOv2 uses randomly
initialised weights because the forward cost is unchanged). This is an explicitly scoped cost
comparison, not a claim that the two branches compute the same function.

\begin{table}[H]
\centering
\caption{Measured front-end latency at $224\times224$. DINOv2 numbers are encoder forward
passes; \dh{} covers spiral sampling, sparse back-projection, and hole taxonomy.}
\label{tab:latency}
\small
\begin{tabular}{llrrr}
\toprule
Branch & device & $p50$ (ms) & learned params & weight size / GPU peak \\
\midrule
DINOv2-L & CPU & 106.5 & 303\,M & 1.2\,GB \\
DINOv2-L & GPU (MPS) & 42.0 & 303\,M & 2.2\,GB peak \\
\textbf{\dh{} core front end} & \textbf{CPU} & \textbf{1.11} & \textbf{0} & \textbf{0 / n.a.} \\
\bottomrule
\end{tabular}
\end{table}

\begin{figure}[H]
\centering
\includegraphics[width=0.62\columnwidth]{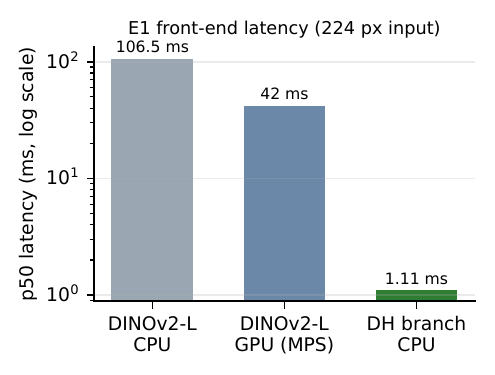}
\caption{Measured front-end latency (log scale). The \dh{} core stages are near-millisecond on
CPU; DINOv2-L is timed as an encoder forward pass.}
\label{fig:latency}
\end{figure}

The measured core front end runs in $1.11$\,ms ($p95\,1.24$, $p99\,1.34$) with OpenCV configured
for 14 CPU threads: $96\times$ lower latency than DINOv2-L on CPU and $38\times$ lower than its
MPS run in this setup, with no learned parameters or weight storage (Fig.~\ref{fig:latency}).
The timing excludes image/depth preprocessing, feature tracking, PnP, and triangulation and
therefore is not an end-to-end pipeline number. At $1440\times1920$ the same measured stages take
$3.81$\,ms. The comparison concerns cost rather than function: DINOv2 emits dense visual
features, while \dh{} emits sparse RGB-D geometry metadata.

\FloatBarrier

\subsection{Held-out triangulation on in-house captures}
\label{sec:inhouse}
On each iPhone capture we split LiDAR-backed trackable points so that PnP pose calibration (set
A) and evaluation (set B) are disjoint, and triangulate B under the A-calibrated pose. A naive
``cyclic'' protocol, which evaluates on the very inliers RANSAC optimised, reports $0.7\%$ error;
the honest hold-out is $5\times$ worse, underscoring why the split matters.

\begin{table}[H]
\centering
\caption{Hold-out triangulation on in-house captures (set B never sees pose
calibration). Camera motion is the dominant factor.}
\label{tab:inhouse}
\small
\begin{tabular}{lrrrr}
\toprule
Capture & hold-out pts & corr & median rel.\ err.\ & reproj.\ \\
\midrule
indoor (near-static) & 225 & 0.963 & 3.5\% & 0.96\,px \\
outdoor (walking) & 2{,}010 & 0.974 & 1.4\% & 0.74\,px \\
\bottomrule
\end{tabular}
\end{table}

Both pass the strict criterion (corr$>0.8$, rel$<10\%$, reproj$<3$px). The walking capture
recovers 663 hole candidates versus 105 indoors (about $6\times$) and is more accurate,
consistent with the method's dependence on translational baseline. Recovered hole candidates in
that capture extend to $6.1$\,m, beyond the useful range observed from the device sensor. When
parallax is small, the gate responds by returning fewer points.

\subsection{Cross-dataset geometry and calibration}
\label{sec:crossdata}
We validate the depth-anchored geometry across scenes on two public benchmarks. On TUM RGB-D
(11 sequences) the scene-median relative error is $\mathbf{2.6\%}$ (per-sequence
$1.0$--$4.2\%$), correlation $0.806$, with $8/11$ passing the strict line. On ARKitScenes
(5 Apple-LiDAR sequences, the closest public analogue to our own hardware) the scene-median
relative error is $\mathbf{6.7\%}$ but correlation drops to $0.745$ and only $1/5$ passes the
strict line---handheld fast motion and $256\times192$ depth make the public set materially
harder. Two honest caveats apply throughout: (i) TUM depth is structured-light, so it validates
the geometry, not iPhone-LiDAR generalisation; and (ii) the medians are computed over points the
gate \emph{keeps}, so low error co-occurs with low coverage (as low as $1\%$ of candidates on a
large-rotation sequence). We make this selection effect explicit rather than hiding it, and the
controlled protocol of \S\ref{sec:e8} fixes the evaluable cohort so that the effect can be
measured rather than hidden. The gate still selects which candidates receive an output.

\paragraph{Calibration.} Binning recovered points by confidence and comparing to actual error
shows that the score generally orders points by risk but is materially miscalibrated. Reliability
bins include both under- and over-confident regions across sequences, with a median expected
calibration error (ECE) of $0.284$ on TUM and $0.224$ on ARKitScenes. An ECE this large is why we
report the score as selective, not calibrated; temperature or isotonic recalibration is left as
future work.

\begin{table}[H]
\centering
\caption{Cross-dataset depth-anchored geometry. Coverage is the fraction of candidates kept;
its variability is the selection effect that \S\ref{sec:e8} measures on a fixed cohort.}
\label{tab:crossdata}
\small
\begin{tabular}{lrrrr}
\toprule
Benchmark & seqs & median rel.\ err.\ & corr & ECE (med.) \\
\midrule
in-house iPhone & 2 & 1.4--3.5\% & 0.96--0.97 & --- \\
TUM RGB-D~\citep{sturm2012tum} & 11 & 2.6\% & 0.806 & 0.284 \\
ARKitScenes~\citep{baruch2021arkitscenes} & 5 & 6.7\% & 0.745 & 0.224 \\
\bottomrule
\end{tabular}
\end{table}

\FloatBarrier

\subsection{Strict near/far range split (controlled far-field recovery)}
\label{sec:e8}
The hold-outs above mix near and far points. To test the core claim directly---\emph{can
near-field returns alone recover true far-field depth?}---we run a controlled protocol on five
ARKitScenes sequences. We mask every LiDAR return beyond a cutoff (default 2\,m), allow PnP to use
\emph{only} the near points, and score recovered far-field depth against the registered
high-resolution \emph{laser} scan, which is independent of the low-res LiDAR. Ground truth is
quality-controlled (nearest-neighbour aligned, $\le 80$\,ms temporal gap; accepted high-res
anchors agree with low-res returns to $1.3\%$ where both exist).

\begin{table}[H]
\centering
\caption{Controlled far-field recovery on ARKitScenes, cutoff $=2$\,m, range-masked cohort
(32{,}690 pooled candidates). Entries are scene medians over five sequences. Triangulation from
the device trajectory is not usable; LiDAR-anchored PnP recovers a majority of evaluable
candidates.}
\label{tab:e8main}
\small
\begin{tabular}{lrrrr}
\toprule
method & coverage & corr & median rel.\ & p90 rel.\ \\
\midrule
sensor-only (after cutoff) & 0.0\% & --- & --- & --- \\
device pose (raw) & 0.4\% & $-0.33$ & 126.0\% & 663\% \\
device pose (GL$\to$CV) & 3.5\% & $-0.44$ & 57.3\% & 375\% \\
\dh{}, no gates & 68.0\% & 0.02 & 14.1\% & 41.0\% \\
\textbf{\dh{}, gated} & \textbf{64.2\%} & \textbf{0.26} & \textbf{13.4\%} & \textbf{35.6\%} \\
\bottomrule
\end{tabular}
\end{table}

\dh{} reaches $\mathbf{64.2\%}$ scene-median coverage of evaluable far-field candidates at
$\mathbf{13.4\%}$ scene-median relative error. Across individual sequences, coverage spans
$47.7$--$88.8\%$ and median relative error $8.5$--$14.1\%$. The pooled output count is
22{,}045 of 32{,}690 candidates ($67.4\%$); we report scene medians in the table so that the
longest sequence does not dominate. Direct use of the device trajectory yields very few, highly
inaccurate outputs, showing that the LiDAR-anchored PnP step---not triangulation alone---provides
usable metric geometry in this test. The gate lowers the scene-median tail (p90
$41.0\!\to\!35.6\%$) at a small coverage cost, while still selecting which candidates receive an
output. At confidence $0.3$, median error falls to $9.5\%$ at $34.3\%$ coverage; stricter
thresholds no longer improve error monotonically.

\begin{figure}[H]
\centering
\includegraphics[width=0.64\columnwidth]{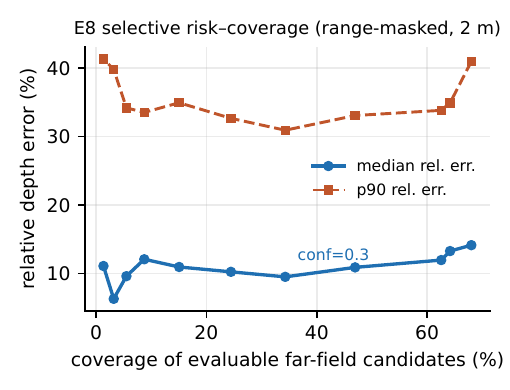}
\caption{Selective risk--coverage on the controlled far-field cohort. Increasing the threshold
initially reduces median and tail error; the knee is near confidence $0.3$, after which stricter
selection no longer improves error monotonically.}
\label{fig:risk}
\end{figure}

\paragraph{Real holes are harder than masked ones.} When we instead score the \emph{true}
low-res LiDAR drop-outs (260 stable GT points after strict filtering), gated \dh{} covers
$53.5\%$ at $42.4\%$ median error and $47.2\%$ p90 error (all scene medians). Without gates the
corresponding values are $58.9\%$, $46.4\%$, and $71.8\%$. Real drop-outs co-occur with edges,
reflections, and occlusion, and are harder than a clean range mask; one of the five sequences
has no gated output on its 12 surviving real-hole candidates. We therefore do \emph{not} claim
high-precision recovery of real holes, and the $6.7\%$ random-hold-out number
(\S\ref{sec:crossdata}) must not be read as a real-hole error. The cutoff sweep
(Table~\ref{tab:e8cutoff}) places the range-masked scene-median error between $11.3$ and $13.4\%$
across $1.5$--$2.5$\,m.

\begin{table}[H]
\centering
\caption{Cutoff sensitivity (range-masked cohort, gated \dh{}). Controlled far-field error is
stable across thresholds; $2$\,m balances sample count and PnP success.}
\label{tab:e8cutoff}
\small
\begin{tabular}{rrrrr}
\toprule
cutoff & PnP success & candidates & coverage & median rel.\ \\
\midrule
1.5\,m & 69.6\% & 113{,}556 & 66.5\% & 11.3\% \\
2.0\,m & 89.2\% & 32{,}690 & 64.2\% & 13.4\% \\
2.5\,m & 91.6\% & 7{,}254 & 71.9\% & 12.9\% \\
\bottomrule
\end{tabular}
\end{table}

\FloatBarrier

\subsection{Effect of the gate}
\label{sec:ablation}
On ARKitScenes we ablate the two gates while sweeping the keyframe gap. Removing the parallax
gate has a severe effect---correlation falls from $0.745$ to $0.034$---whereas removing the
reprojection gate is mild ($0.745\!\to\!0.724$). Across $16$--$60$-frame gaps the relative error
stays within $6.3$--$7.1\%$ while coverage and correlation trade off against parallax. The
parallax gate is thus the core safety mechanism: it is what implements abstention, directly
supporting the design of \S\ref{sec:gate}.

\begin{table}[H]
\centering
\caption{Gate ablation on ARKitScenes (scene-median). The parallax gate is essential.}
\label{tab:gateablation}
\small
\begin{tabular}{lrr}
\toprule
configuration & corr & median rel.\ err.\ \\
\midrule
gap 16 & 0.803 & 7.1\% \\
gap 30 (default) & 0.745 & 6.7\% \\
gap 60 & 0.731 & 6.3\% \\
parallax gate OFF (gap 30) & 0.034 & 7.2\% \\
reprojection gate OFF (gap 30) & 0.724 & 6.8\% \\
\bottomrule
\end{tabular}
\end{table}

\FloatBarrier

\subsection{Comparison to learning-based multi-view geometry}
\label{sec:baselines}
We run DUSt3R~\citep{wang2024dust3r} and VGGT-1B~\citep{wang2025vggt} on the same five
ARKitScenes sequences. The input conditions are \emph{not} equal and we state this plainly: the
learned baselines are RGB-only and receive a \emph{per-frame ground-truth oracle scale} (fit on a
disjoint pixel set), whereas \dh{} is LiDAR-anchored and metric with \emph{no} scale alignment.

\begin{table}[H]
\centering
\caption{Reference results on ARKitScenes under different inputs and scale protocols. The learned
baselines receive per-frame ground-truth oracle scale; \dh{} receives sparse LiDAR and no
post-hoc scale alignment. These are contextual measurements, not a head-to-head ranking.}
\label{tab:baselines}
\small
\begin{tabular}{lrrccl}
\toprule
method & median rel.\ & corr & GT scale? & metric? & compute \\
\midrule
\textbf{\dh{}} & 6.7\% & 0.745 & no & yes & 0-param, CPU real-time \\
DUSt3R & 7.2\% & 0.802 & oracle & no & 571\,M, GPU, 78\,s/vid \\
VGGT-1B & \textbf{2.5\%} & \textbf{0.967} & oracle & no & 1.26\,B, GPU, 235\,s/vid \\
\bottomrule
\end{tabular}
\end{table}

VGGT-1B is clearly the most accurate after scale alignment. DUSt3R's median error is numerically
close to \dh{}, but the inputs, output density, and scale treatment differ, so no equivalence or
superiority follows from that proximity. Among the three implementations evaluated here, \dh{}
is the one that emits metric depth without post-hoc alignment, with explicit abstention and zero
learned parameters.

\FloatBarrier

\subsection{Downstream behaviour as a visual front-end}
\label{sec:downstream}
Finally we place the geometry features in two downstream proxy regressions. On RH20T (799 tasks,
$\sim$100k samples, task-hold-out, ridge probe over a shared robot-context baseline), future
end-effector-motion regression with \dh{} features has validation MSE within $0.6$--$1.3\%$ of
DINOv2-L (ratio $1.006$ over all samples and $1.013$ on moving samples). This is a descriptive
difference; we did not run an equivalence or non-inferiority test. On a single-scene
visual-odometry proxy, \dh{} gives validation MSE $1.32$ versus $1.63$ for DINOv2 (ratio $0.81$).
The latter is a geometry-favoured, single-scene result, and the broader proxy is only weakly
visual: every front end improves on context alone by $0.6$--$2.4\%$, and \dh{} does not beat raw
RGB at scale. These experiments show that the features can be consumed by the probes at similar
error, not that they preserve policy performance or improve action. That question requires a
visually demanding task and a closed-loop evaluation.

\section{Limitations}
\label{sec:limitations}
Our claims are bounded, by design. \textbf{No robot in the loop:} all downstream evidence is
offline regression, never a task success rate; the RH20T comparison is descriptive and includes
no equivalence test. \textbf{Scoped latency:} the $1.11$\,ms measurement covers the core
single-frame stages, not tracking, PnP, triangulation, or preprocessing. \textbf{Map
fragmentation:} sparse single-view spiral structure does not close
into a room; triangulation widens coverage but does not stitch fragments, an orthogonal problem
we do not solve. \textbf{Motion dependence:} recovery scales with translational baseline; nearly
static hand-held clips give too little parallax and the gate (correctly) abstains. \textbf{Public
data is harder:} on ARKitScenes only $1/5$ sequences pass the strict correlation line, even as
relative error stays $\sim$5--9\%; we do not claim iPhone-LiDAR generalisation from
structured-light TUM. \textbf{Selective, not calibrated:} ECE is $\sim$0.22--0.28 and the
risk--coverage curve is not strictly monotone; the score ranks but is not a probability.
\textbf{Real holes:} the most direct measurement (\S\ref{sec:e8}) puts real-hole recovery at
$42.4\%$ median error on a small 260-point GT set, with one sequence yielding no gated output.
This is not a precision-depth result. We state these limits so that no number in this paper is
read beyond what it supports.

\section{Conclusion}
We presented \dh{}, a training-free geometry back-end that treats a short-range LiDAR not as the
sole depth source but as a metric ruler: near returns anchor the two-view pose, that pose
triangulates the far field, and a parallax/reprojection gate abstains where the geometry is weak,
attaching a selective confidence to what remains. Under a strict controlled-cutoff protocol with
independent laser ground truth, the method reaches $64.2\%$ scene-median coverage of evaluable
spiral-sampled far-field candidates at $13.4\%$ scene-median relative error using only sub-2\,m
returns; direct triangulation from the device trajectory is not usable. Its measured core front
end runs in $1.11$\,ms on CPU with zero learned parameters. It does not match a billion-parameter
learned model after oracle scale alignment, nor is that its aim. Alongside the positive results,
we report the falsified
alternatives---defocus, focus stacks, fusion, ICP-over-good-VIO, attention-to-holes---because they
draw the boundary of where this honest, lightweight geometry is the right tool. We hope both halves
are useful to anyone building edge perception that must say not only what it sees, but where it
cannot see.

\paragraph{Reproducibility.} Complete source, unit tests, and per-experiment configuration/result
logs are prepared for public release alongside the paper. Public datasets are obtained from their
official sources; no data or third-party weights are redistributed.

\small
\bibliographystyle{plainnat}
\bibliography{references}

\end{document}